\documentclass[letterpaper, 10 pt, conference]{ieeeconf}  
\IEEEoverridecommandlockouts                              
\makeatletter
\let\NAT@parse\undefined
\makeatother

\usepackage{amsmath,amsfonts}
\usepackage{algorithmic}
\usepackage{algorithm}
\usepackage{array}
\usepackage{makecell}
\usepackage[caption=false,font=normalsize,labelfont=sf,textfont=sf]{subfig}

\usepackage{textcomp}
\usepackage{stfloats}
\usepackage{url}
\usepackage{verbatim}
\usepackage{graphicx}

\usepackage{multirow} 
\usepackage{pifont} 
\usepackage{colortbl} 
\usepackage{xcolor}

\usepackage[
  colorlinks=true,
  linkcolor=cyan,
  urlcolor=black,
  citecolor=green
]{hyperref}

\title{\textbf{Decision-Driven Semantic Object Exploration for Legged Robots via Confidence-Calibrated Perception and Topological Subgoal Selection}}

\author{Guoyang Zhao$^{1}$, Yudong Li$^{2}$, Weiqing Qi$^{1}$, Kai Zhang$^{1}$, Bonan Liu$^{1}$, Kai Chen$^{1}$, Haoang Li$^{1}$ and Jun Ma$^{1}$
\thanks{$^{1}$Guoyang Zhao, Weiqing Qi, Kai Zhang, Bonan Liu, Kai Chen, Haoang Li and Jun Ma are with the Robotics and Autonomous Systems Thrust, The Hong Kong University of Science and Technology (Guangzhou), China
(email: {\{gzhao492, wqiad, kzhang740, bliu404, kchen916\}@connect.hkust-gz.edu.cn; haoangli@hkust-gz.edu.cn; jun.ma@ust.hk})}
\thanks{$^{2}$Yudong Li is with the Department of Mechanical and Energy Engineering, Southern University of Science and Technology, China
(email:{12432385@mail.sustech.edu.cn})}
}

\begin{document}

\maketitle
\thispagestyle{empty}
\pagestyle{empty}

\begin{abstract}
Conventional navigation pipelines for legged robots remain largely geometry-centric, relying on dense SLAM representations that are fragile under rapid motion and offer limited support for semantic decision making in open-world exploration.
In this work, we focus on decision-driven semantic object exploration, where the primary challenge is not map consistency but how noisy and heterogeneous semantic observations can be transformed into stable and executable exploration decisions.
We propose a vision-based approach that explicitly addresses this problem through confidence-calibrated semantic evidence arbitration, a controlled-growth semantic topological memory, and a semantic utility-driven subgoal selection mechanism.
These components enable the robot to accumulate task-relevant semantic knowledge over time and select exploration targets that balance semantic relevance, reliability, and reachability, without requiring dense geometric reconstruction.
Extensive experiments in both simulation and real-world environments demonstrate that the proposed mechanisms consistently improve the quality of semantic decision inputs, subgoal selection accuracy, and overall exploration performance on legged robots.
\end{abstract}

\section{Introduction}

Autonomous exploration and navigation in open, unstructured environments remain key challenges for mobile robots. These capabilities are essential for applications such as search-and-rescue, warehouse logistics, and environmental monitoring, where robots must explore unknown spaces, interpret scene semantics, and search for task-relevant objects \cite{shah2023lm}. Compared with wheeled platforms, legged robots offer superior terrain adaptability, but their rapid motion, frequent ground impacts, and viewpoint instability significantly complicate perception and decision making.

Most existing navigation systems rely on SLAM-based pipelines that perform geometric mapping and pose optimization using LiDAR or multi-sensor fusion to construct dense metric maps \cite{placed2023survey}. Although effective in structured environments, such geometry-driven approaches require expensive sensors, precise calibration, and substantial computational resources, limiting their deployment on lightweight or cost-sensitive platforms that primarily rely on cameras \cite{liu2024omnicolor}. Moreover, for task-driven semantic object exploration, constructing a globally consistent dense map is often unnecessary, as it does not directly address the key decision of where to explore next or which semantic targets to pursue.

\begin{figure}[t]
    \centering
    \includegraphics[width=.48\textwidth]{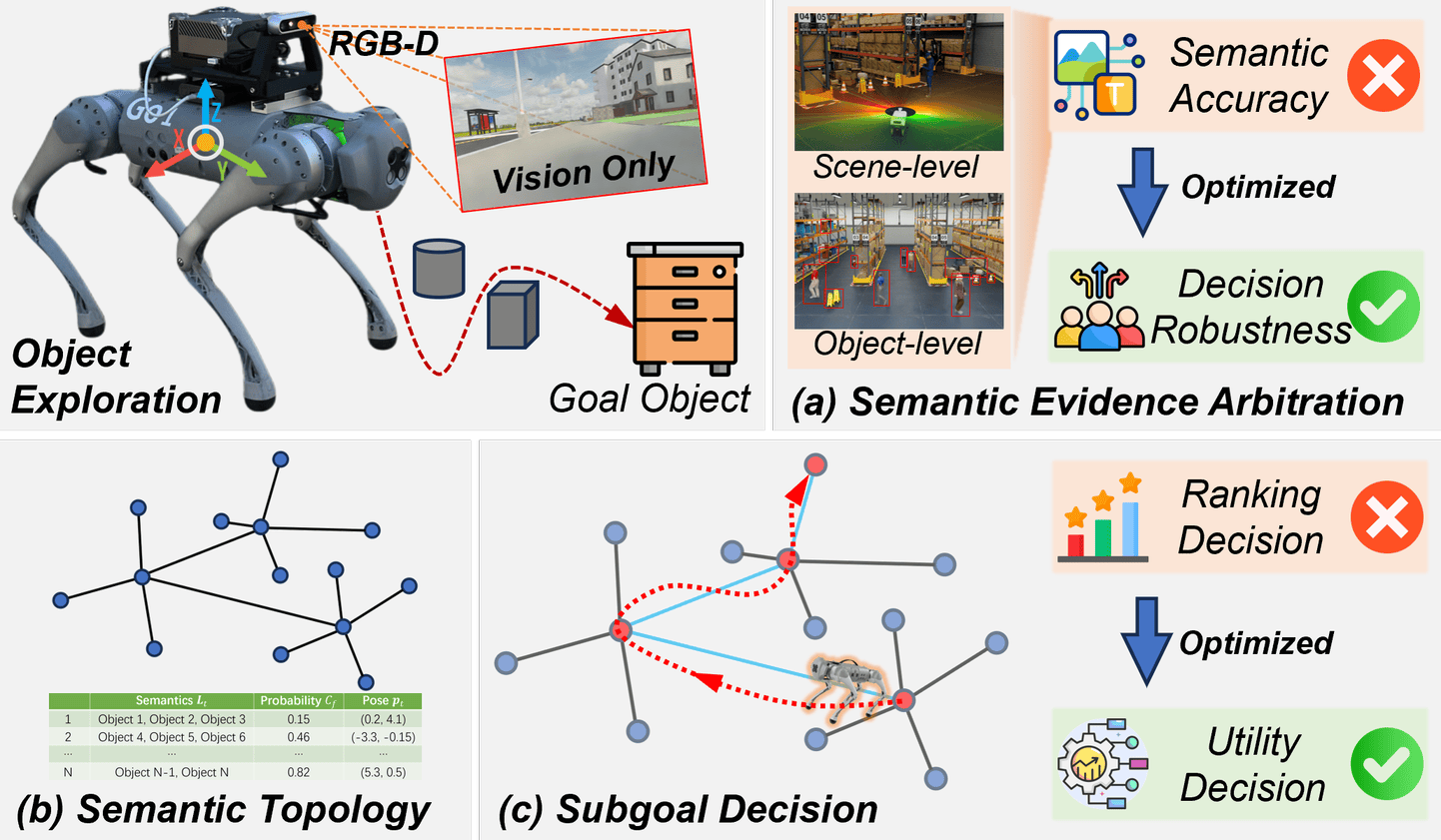}
    \vspace{-19pt}
\caption{\textbf{Decision-driven vision-only semantic object exploration for legged robots.}
(a) Semantic evidence arbitration calibrates scene- and object-level perception to produce decision-robust semantic targets.
(b) A semantic topology represents explored locations as nodes enriched with semantic cues.
(c) Subgoals are selected via utility-driven decision making rather than simple ranking.}
    \label{cover-figure}
    \vspace{-18pt}
\end{figure}

From a task perspective, open-world semantic object exploration is fundamentally a decision-making problem rather than purely a mapping or localization task \cite{huang2023visual}. A robot must reason under partial and uncertain observations to determine which semantic hypotheses are reliable \cite{zhao2025fisheyedepth}, which regions are likely to contain the target object, and how to select actions that balance exploration efficiency and execution cost. Compared with dense geometric maps, topological representations offer a compact abstraction of spatial connectivity and naturally support semantic integration, making them more suitable for long-horizon, goal-directed exploration.

Recent advances in Vision-Language Models (VLMs) and Vision-Language Navigation (VLN) have enabled robots to infer rich semantic cues from visual observations and natural language instructions \cite{yokoyama2024vlfm}. These models demonstrate strong potential for open-vocabulary object understanding and semantic reasoning. However, most existing approaches focus primarily on perception or recognition, with limited attention to explicit decision modeling, closed-loop execution, and fine-grained obstacle avoidance \cite{roth2024viplanner}. In addition, many methods are evaluated mainly in simulation or static settings, leaving open questions regarding their deployment on real-world, heterogeneous legged robot platforms.

In this work, we revisit semantic object exploration from a decision-driven perspective. Rather than emphasizing dense geometric reconstruction, we focus on how semantic information can be transformed into actionable decisions under camera-centric sensing and constrained computation. We propose a decision-driven semantic object exploration approach that combines confidence-calibrated visual perception with topology-level subgoal selection. By maintaining a compact semantic topological memory and selecting subgoals based on semantic utility, our method enables efficient and interpretable exploration without requiring dense metric maps.
Our main contributions are summarized as follows:
\begin{itemize}
    \item We propose a confidence-calibrated semantic target arbitration mechanism that integrates scene-level and object-level visual cues to produce reliable, executable exploration targets under partial observations.
    \item We introduce a controlled-growth semantic topological memory that compactly represents exploration history and supports long-horizon semantic decision making.
    \item We design a semantic utility-driven subgoal selection strategy that jointly considers semantic relevance, confidence, and travel cost, enabling real-time execution.
    \item We validate the proposed approach through extensive experiments on multiple simulated legged robot platforms and a real Go1 robot, demonstrating cross-platform deployability and practical feasibility.
\end{itemize}


\section{Related Work}
\label{sec:related}

\subsection{Semantic Evidence for Open-World Exploration}

Recent advances in VLMs have improved semantic understanding in open-world environments. Models such as CLIP~\cite{radford2021learning}, BLIP-2~\cite{li2023blip}, and Qwen-VL~\cite{bai2023qwen} leverage large-scale vision-text pretraining to achieve strong zero-shot recognition, image captioning, and visual question answering. These scene-level models provide holistic semantic descriptions and strong generalization, but lack precise spatial localization and often fail to distinguish specific target instances, limiting their use in robotic navigation.

To address this, object-level semantic perception models have been developed. Methods such as the Segment Anything Model~\cite{kirillov2023segment}, Grounding DINO~\cite{liu2023grounding}, and YOLO-World~\cite{cheng2024yolo} enable promptable segmentation or open-vocabulary object detection, allowing recognition and localization of unseen object categories. However, they typically lack global contextual awareness and treat objects as isolated instances without modeling relationships with the surrounding scene.

Overall, existing approaches provide semantic evidence at different granularities, but most treat these outputs purely as perception results. How to model the uncertainty and complementarity of scene-level and object-level cues, and calibrate them for downstream decision making in exploration tasks, remains underexplored.

\subsection{Semantic Memory and Environment Abstraction}

Environment representation and memory are fundamental to robotic exploration. 
Traditional approaches rely on SLAM-based systems to build dense geometric maps through feature matching and optimization. 
Representative examples include visual-inertial systems such as ORB-SLAM3~\cite{campos2021orb} and LiDAR-inertial odometry methods~\cite{xu2022fast2}. 
While effective in structured environments, these methods require precise calibration and substantial computation, and their robustness degrades in dynamic or large-scale open-world scenarios. 
Moreover, purely geometric maps lack semantic information, limiting their utility for language-guided exploration and task-level reasoning.

To reduce mapping complexity and improve task relevance, recent work has explored lightweight or geometry-agnostic representations. 
One line of research learns reactive navigation policies through reinforcement or imitation learning without explicit map construction~\cite{zhu2017target,pfeiffer2018reinforced}. 
Although efficient, such approaches often suffer from limited interpretability and generalization. 
Another line adopts topological or graph-based representations, where nodes represent semantic landmarks or regions and edges encode reachability or transition costs~\cite{chaplot2020neural}. 
These representations provide compact environment abstractions and naturally support semantic integration for long-horizon exploration.

Despite these advances, existing semantic or topological representations largely focus on geometric connectivity, with limited modeling of semantic uncertainty, dynamic updates, or exploration-aware management. 
Consequently, their capability to support decision-oriented semantic exploration in open-world environments remains limited.

\subsection{Decision Making for Semantic Object Exploration}

Building on semantic perception and environment representation, decision making is central to effective object exploration. VLN and related tasks incorporate natural language to guide robots toward semantic goals. Representative approaches include VL-Nav~\cite{du2025vlnav}, HOV-SG \cite{werby23hovsg}, VLFM \cite{yokoyama2024vlfm}, and cross-modal map learning methods~\cite{georgakis2022cross}, which combine visual, linguistic, and spatial representations for navigation. Additional works explore volumetric or structured representations to enhance scene understanding~\cite{liu2024volumetric}.

While these methods advance the integration of language and vision, their decision processes are often implicit, relying on heuristic rules or end-to-end policies for subgoal selection. Moreover, most evaluations are conducted in simulation or relatively static environments, leaving open questions regarding real-world deployment on mobile robots, especially legged platforms subject to dynamic motion, obstacle avoidance, and real-time constraints.

Consequently, how heterogeneous and uncertain semantic evidence can be transformed into explicit, interpretable, and real-time decision rules for long-horizon semantic object exploration remains an open research problem.

\section{Methodology}

\subsection{Task Definition and Problem Formulation}
\label{sec:task_formulation}

We study \emph{semantic object exploration} in open-world environments, where a robot searches for task-relevant targets in previously unseen spaces given a natural language instruction $\ell$ specifying an open-vocabulary object category or description. 
Rather than maintaining a globally consistent dense metric map, the robot performs \emph{decision-oriented exploration} by sequentially selecting executable semantic subgoals under partial observability until the target is found and reached.

At time $t$, the robot receives observations $\mathbf{o}_t=\{I_t,D_t\}$, where $I_t$ and $D_t$ denote the RGB and depth images, respectively, together with proprioceptive signals $\mathbf{z}_t$ (odometry and IMU). 
The instruction $\ell$ constrains task-relevant semantics, while the robot gathers \emph{heterogeneous} evidence from scene-level vision-language reasoning and object-level open-vocabulary detection. 
Scene-level cues provide global context and directional guidance, whereas object-level detections yield spatially grounded candidates but may be noisy under motion and occlusion. 
We formulate exploration as a sequential decision process in which the robot maintains an internal memory state $m_t$ and selects an executable subgoal $g_t$ at each step to achieve efficient task completion while satisfying real-time safety constraints.

\begin{figure*}[t!]
    \centering
    \includegraphics[width=.99\textwidth]{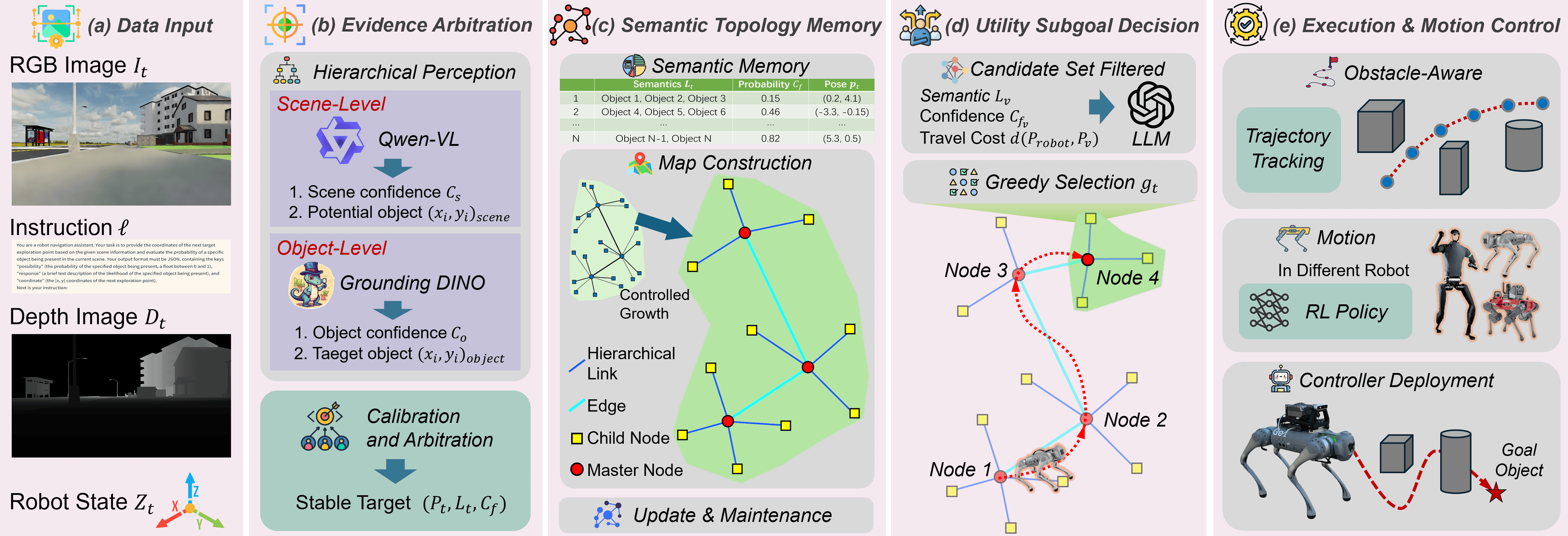}
    \vspace{-5pt}
\caption{\textbf{Overview of the decision-driven semantic object exploration framework.}
(a) RGB image $I_t$, depth image $D_t$, instruction $\ell$, and robot state $Z_t$ are used as inputs. (b) Hierarchical perception extracts semantic evidence, which is calibrated and fused to produce a stable target $(P_t, L_t, C_f)$. (c) Explored regions are organized into a semantic topological memory with map construction and maintenance. (d) Candidate nodes are filtered and evaluated to select the next subgoal. (e) Obstacle-aware trajectory tracking and RL-based motion policies enable execution on different robot platforms.}
    \label{framework}
    \vspace{-15pt}
\end{figure*}



\subsection{Overview of Exploration Framework}
\label{sec:framework_overview}
As illustrated in Fig.~\ref{framework}, exploration is organized as an 
\emph{evidence-memory-decision-execution} pipeline. 
Given RGB-D observations and the instruction $\ell$, the system extracts semantic evidence from the current view and integrates it into a semantic topological memory encoding spatial connectivity, semantic labels, and confidence cues. 
A decision module ranks candidate nodes using a semantic utility criterion, while a local obstacle-aware planner generates short-horizon motion commands for execution. 
The framework centers on three components: evidence arbitration, controlled-growth semantic memory, and utility-driven subgoal decision.

At the beginning of each episode, the robot performs a brief panoramic observation by rotating in place to collect multi-view RGB-D frames and state estimates, initializing the semantic topological memory with a root node and initial semantic candidates. 
The system then iterates evidence extraction, memory update, subgoal selection, and motion execution during exploration.

\subsection{Confidence-Calibrated Semantic Evidence Arbitration}
\label{sec:evidence_arbitration}

In open-world exploration, the robot must extract task-relevant \emph{semantic evidence} from the current view and convert it into executable target candidates.
Scene-level vision-language reasoning and object-level open-vocabulary detection provide complementary evidence: the former captures global context and directional cues, while the latter yields spatially grounded candidates.
However, their confidence behaviors and spatial reliability differ under motion and occlusion.
We therefore introduce a semantic evidence arbitration mechanism that integrates heterogeneous evidence on a unified posterior scale with lightweight spatial-consistency and feasibility constraints, producing stable semantic targets for downstream memory update and decision making.

\subsubsection{\textbf{Scene-level Semantic Evidence}}
Given the RGB input $I_t$ and instruction $\ell$, we query Qwen2.5-VL-7B~\cite{bai2023qwen} to obtain a set of scene-level proposals
$\{(\mathbf{p}^{s}_i, L^{s}_i)\}_{i=1}^{N_s}$,
where $\mathbf{p}^{s}_i\in\mathbb{R}^2$ denotes an image-plane candidate location and $L^{s}_i$ is the associated semantic label.
The model also provides a scene-level confidence $C_s\in[0,1]$ that reflects the reliability of the global interpretation for the current view.

\subsubsection{\textbf{Object-level Semantic Evidence}}
For open-vocabulary detection, we adopt GroundingDINO-T~\cite{liu2023grounding}, producing a set of detections
$\{(\mathbf{b}_j, L^{o}_j, C_o(j))\}_{j=1}^{N_o}$,
where $\mathbf{b}_j\in\mathbb{R}^4$ is the bounding box, $L^{o}_j$ is the text-aligned label, and $C_o(j)\in[0,1]$ is the detection confidence for the $j$-th box.
We use the box center $\mathbf{p}^{o}_j\in\mathbb{R}^2$ as an object-level localization cue.

\subsubsection{\textbf{Calibration and Arbitration}}
Scene- and object-level evidence are mapped onto a unified posterior scale.
To suppress low-confidence noise, we apply a monotonic confidence calibration to any confidence value $C\in[0,1]$:
\begin{equation}
\tilde{C}=\min\!\left(1,\max\!\left(0,\frac{C-\tau}{1-\tau}\right)\right),
\label{eq:conf_calib}
\end{equation}
where $\tau\in[0,1)$ is a fixed threshold.
Applying Eq.~\eqref{eq:conf_calib} to $C_s$ and $C_o(j)$ yields calibrated confidences $\tilde{C}_s$ and $\tilde{C}_o(j)$, respectively.

Candidate targets consist of (i) object detections aligned with the instruction $\ell$, and (ii) scene proposals indicating unexplored directions.
When both sources are available, we associate a scene proposal with an object detection by maximum spatial overlap.
For each candidate target $t$, we denote its associated object box by $\mathbf{b}_t$, its associated scene region by $r^{s}_t$, and its calibrated confidences by $\tilde{C}_o(t)$ and $\tilde{C}_s(t)$ (if one source is missing, the corresponding term is omitted).
We then define a posterior score:
\begin{equation}
S(t) \propto P(t)\, \tilde{C}_o(t)^{\beta_o}\, \tilde{C}_s(t)^{\beta_s}\,
\big(\mathrm{IoU}(r^{s}_t,\mathbf{b}_t)+\epsilon\big)^{\lambda}\,
\big(\kappa_t+\epsilon\big)^{\mu},
\label{eq:bayes_fusion}
\end{equation}
where $P(t)$ is an optional prior, $\mathrm{IoU}(\cdot)\in[0,1]$ enforces spatial consistency between the scene region $r^{s}_t$ and the object box $\mathbf{b}_t$, and $\kappa_t\in[0,1]$ is a depth-based feasibility indicator that suppresses unreachable targets.
$\beta_o,\beta_s,\lambda,\mu$ are fixed hyperparameters controlling relative contributions, and $\epsilon\approx 10^{-3}$ ensures numerical stability.
The final target is selected by maximum posterior inference:
\begin{equation}
t^{*}=\arg\max_{t} S(t),\qquad
C_f=\frac{S(t^{*})}{\sum_{t} S(t)},
\label{eq:posterior_score}
\end{equation}
where $C_f\in[0,1]$ is the normalized fused confidence of the selected target.

Let $\mathbf{p}^{o}_t$ denote the object-level location (box center) associated with $t$ and $\mathbf{p}^{s}_t$ denote the scene-level proposal location associated with $t$.
If both are available, we compute the executable target location by confidence-weighted interpolation:
\vspace{-3pt}
\begin{equation}
\mathbf{p}_t=w\,\mathbf{p}^{o}_t+(1-w)\,\mathbf{p}^{s}_t,\quad
w=\frac{\tilde{C}_o(t)^{\beta_o}}{\tilde{C}_o(t)^{\beta_o}+\tilde{C}_s(t)^{\beta_s}}.
\label{eq:location_fusion}
\end{equation}
Otherwise, the available location is used directly as $\mathbf{p}_t$.
The module outputs $(\mathbf{p}_t, L_t, C_f)$, where $L_t$ denotes the semantic label of the selected target, for downstream memory update and subgoal decision.

\subsection{Controlled-Growth Semantic Topological Memory}
\label{sec:topo_memory}

To support long-horizon semantic object exploration, we maintain a \emph{controlled-growth semantic topological memory} that encodes explored locations, semantic evidence reliability, and remaining exploration opportunities.
The memory is represented as a graph $G=(V,E)$, where each node corresponds to a revisitable location and each edge represents traversable connectivity with an associated cost.

\subsubsection{\textbf{Graph Structure and Node State}}
Each node $v_i\in V$ stores its 3D position $\mathbf{P}_{v_i}\in\mathbb{R}^3$, semantic label $L_{v_i}$, fused confidence $C_{f_{v_i}}$ from the evidence arbitration module (Sec.~\ref{sec:evidence_arbitration}), and an exploration potential $P_{\text{explore}}(v_i)\in[0,1]$ indicating remaining exploration value.
The initial viewpoint is designated as the \emph{master node} $v_m$.
Given a target pixel $\mathbf{p}_t$ and its depth $D_t(\mathbf{p}_t)$ in the current frame, we back-project it to a 3D node candidate in the world frame:
\begin{equation}
\mathbf{P}_{v_c}=\mathbf{T}_{w\leftarrow c}(t)\,\pi^{-1}\!\big(\mathbf{p}_t, D_t(\mathbf{p}_t)\big),
\label{eq:backproject_node}
\end{equation}
where $\pi^{-1}(\cdot)$ denotes the camera back-projection using intrinsics, and $\mathbf{T}_{w\leftarrow c}(t)$ is the camera-to-world transform at time $t$ obtained from odometry and IMU without global pose-graph optimization.

\subsubsection{\textbf{Controlled Growth and Connectivity}}
To bound memory growth, we apply a distance- and confidence-gated update rule.
Let $v_{\text{nn}}$ denote the nearest existing node to $\mathbf{P}_{v_c}$ in Euclidean distance.
A new node is inserted only if $\|\mathbf{P}_{v_c}-\mathbf{P}_{v_{\text{nn}}}\|_2>\delta_d$ and the fused confidence $C_f$ exceeds a fixed threshold.
Otherwise, the observation is merged into $v_{\text{nn}}$ by updating its attributes and refreshing confidence via an exponential moving average:
\begin{equation}
C_{f_{v_{\text{nn}}}} \leftarrow (1-\rho)\,C_{f_{v_{\text{nn}}}} + \rho\,C_f,
\label{eq:cf_merge}
\end{equation}
where $\rho\in(0,1]$ is a fixed smoothing factor shared across experiments.

When a new node is inserted, edges are added to spatial neighbors within radius $\delta_r$ or to the $k$ nearest nodes to maintain connectivity.
Traversal costs on edges are approximated by Euclidean distance or a short-horizon reachable cost, yielding a compact yet connected graph during exploration.

\subsubsection{\textbf{Exploration Potential Update and Pruning}}
The exploration potential $P_{\text{explore}}(v_i)$ is initialized from fused confidence and decays when the node is revisited or its surroundings are observed:
\begin{equation}
P_{\text{explore}}(v_i) \leftarrow (1-\xi)\,P_{\text{explore}}(v_i),
\label{eq:explore_decay}
\end{equation}
where $\xi\in(0,1]$ is a fixed decay rate.

To further limit memory size, we periodically prune nodes with both low exploration potential and low semantic confidence, while spatially adjacent nodes with consistent semantics may be merged.
This maintenance keeps the semantic topological memory compact and stable for long-horizon subgoal decision.

\subsection{Semantic Utility-Driven Subgoal Decision}
\label{sec:utility_decision}

Given the semantic topological memory $G=(V,E)$ (Sec.~\ref{sec:topo_memory}), the robot converts accumulated semantic evidence and memory states into the next executable subgoal.
We rank candidate nodes using a unified \emph{semantic utility} that jointly considers instruction relevance, evidence reliability, exploration value, and reachability cost.

\subsubsection{\textbf{Candidate Set and Semantic Relevance}}
Let $\mathcal{V}_c\subseteq V$ denote the candidate set filtered by exploration potential:
\vspace{-3pt}
\begin{equation}
\mathcal{V}_c=\{v\in V \mid P_{\text{explore}}(v)\ge \tau_p\},
\label{eq:candidate_set}
\end{equation}
where $\tau_p\in[0,1]$ is a fixed threshold.
For each candidate $v\in\mathcal{V}_c$, we retrieve its semantic label $L_v$, fused confidence $C_{f_v}$, and travel cost $d(\mathbf{P}_{\text{robot}},\mathbf{P}_v)$, approximated by the shortest-path cost on $G$ or Euclidean distance.
To evaluate semantic relevance between the instruction $\ell$ and node semantics $L_v$, we employ a large language model (LLM) \cite{achiam2023gpt} as a lightweight adjudicator producing a normalized relevance score $S^{\text{LLM}}_{v}\in[0,1]$.
For reproducibility, we use a fixed prompt template and deterministic decoding (temperature $=0$), and the LLM only performs high-level ranking.

\subsubsection{\textbf{Semantic Utility and Greedy Selection}}
The semantic utility of a candidate node $v\in\mathcal{V}_c$ is defined as
\vspace{-3pt}
\begin{equation}
U(v)=\Big(S^{\text{LLM}}_{v}\Big)^{\alpha}\cdot
\Big(C_{f_v}\Big)^{\beta}\cdot
\Big(P_{\text{explore}}(v)\Big)^{\eta}\cdot
\exp\!\big(-\gamma\, d(\mathbf{P}_{\text{robot}},\mathbf{P}_v)\big),
\label{eq:semantic_utility}
\end{equation}
where $\alpha,\beta,\eta$ weight semantic relevance, confidence stability, and exploration potential, respectively, and $\gamma$ balances semantic gain against travel cost.
The next subgoal is selected greedily as
\vspace{-3pt}
\begin{equation}
v^*=\arg\max_{v\in\mathcal{V}_c} U(v), 
\qquad g_t \leftarrow \mathbf{P}_{v^*},
\label{eq:greedy_subgoal}
\end{equation}
where $v^*$ is the selected node and $g_t\in\mathbb{R}^3$ is the corresponding 3D subgoal position for downstream local planning and execution.
This rule prioritizes nodes that are semantically relevant, evidence-supported, exploration-informative, and easy to reach, while reducing oscillations caused by noisy semantic cues.

\subsection{Execution Interface and Motion Realization}
\label{sec:execution_motion}

The subgoal $g_t$ selected by the semantic utility module (Sec.~\ref{sec:utility_decision}) must be translated into safe and executable actions on real legged robots.
We decouple high-level semantic decision making from low-level dynamics using a two-stage execution interface: obstacle-aware command generation followed by policy-based motion realization.

\subsubsection{\textbf{Local Obstacle-aware Command Generation}}
Given a 3D subgoal $\mathbf{P}_{g_t}$, we project it into the current view to obtain $(x_{g_t},y_{g_t})$ and compute short-horizon collision-avoiding commands from the depth observation $D_t$.
We employ \textit{Viplanner}~\cite{roth2024viplanner}, an end-to-end vision-based local planner, to generate linear and angular velocities
$
[v_p,\omega_p]^T = f_{\text{viplanner}}(D_t, x_{g_t}, y_{g_t}),
$
which guide the robot toward the subgoal while avoiding nearby obstacles.

\subsubsection{\textbf{Policy-based Motion Realization}}
For deployment on heterogeneous legged platforms, we use reinforcement-learning-based locomotion policies as low-level controllers.
Given the robot state $s_t$ and command $[v_p,\omega_p]^T$, the policy outputs joint-level actions
$
\mathbf{u}_t = \pi_{\theta}(s_t, v_p, \omega_p).
$
A lightweight interface wrapper incorporates recent execution history $\mathcal{H}_{t-1}$ (e.g., past actions and contact states) to stabilize motion during contact transitions.
This execution interface preserves the modularity of the navigation stack: high-level semantic exploration remains morphology-agnostic, while motion realization ensures smooth and safe behaviors on real hardware.

\section{Experiments}
\subsection{Experimental Setup}
Simulation experiments and local GroundingDINO inference are conducted on a workstation equipped with an NVIDIA RTX4090 GPU. GPT-4 and Qwen-VL are accessed via their official cloud APIs for remote inference.
In real-world experiments, a Jetson AGX Orin onboard the robot serves as the edge computing unit, responsible for planning and control. The workstation and robot communicate wirelessly through ROS, enabling real-time exchange of image streams and subgoal commands.

\subsubsection{\textbf{Simulation Environment}}
We instantiate two environments in Isaac Sim: an outdoor urban scene (\emph{CARLA Town}) and an indoor scene (\emph{Warehouse}), within which five semantic exploration tasks are defined. Each task is executed from randomized initial poses with fixed random seeds to ensure reproducibility. To evaluate cross-morphology generalization, we further perform experiments under identical settings using additional quadruped and humanoid robot models.

\subsubsection{\textbf{Real-World Environment}}
Real-world validation is conducted on a Unitree Go1 quadruped robot equipped with a RealSense D435i RGB-D camera.
Experiments are performed in five environments: an office (30 m$^2$) with randomly placed desks, chairs, and cabinets; a showroom (40 m$^2$) containing display boards and machines; a laboratory (30 m$^2$) with manually constructed obstacle courses; a living room (30 m$^2$) furnished with household objects; and an outdoor garden (50 m$^2$) featuring grass and tree obstacles.
RGB-D streams are recorded at 30 FPS, while pose and odometry are obtained from the Go1 internal state estimator.

\subsubsection{\textbf{Evaluation Metrics}}
We evaluate the proposed decision mechanisms and downstream navigation performance using a unified set of metrics.
\textbf{(i) Evidence and decision evaluation.}
Semantic evidence quality is measured by the \textit{Semantic Accuracy (SA)}, defined as
$\mathrm{SA}=\frac{1}{N}\sum_{i=1}^{N}\mathbb{I}\!\left[\hat{c}_i=c_i\right]$,
where $N$ is the number of trials, $\hat{c}_i$ and $c_i$ denote the reached and instructed categories in trial $i$, and $\mathbb{I}[\cdot]$ is the indicator function.
Global decision quality on the topological memory is quantified by the \textit{Global Node Selection Accuracy (GNSA)},
$\mathrm{GNSA}=\frac{1}{\sum_{i=1}^{N} T_i}\sum_{i=1}^{N}\sum_{t=1}^{T_i}\mathbb{I}\!\left[v^{*}_{i,t}=v^{\mathrm{oracle}}_{i,t}\right]$,
where $T_i$ is the number of decision steps in episode $i$, $v^{*}_{i,t}$ is the selected node, and $v^{\mathrm{oracle}}_{i,t}$ is the oracle node.
\textbf{(ii) Navigation evaluation.}
Navigation safety is quantified by the \textit{Obstacle Avoidance Success Rate (OASR)}, defined as
$\mathrm{OASR}=1-\frac{1}{N}\sum_{i=1}^{N}\mathbb{I}[c_i]$,
where $c_i$ indicates a collision event in trial $i$ (i.e., breaching the predefined clearance threshold).
Overall performance is measured by \textit{Success Rate (SR)} and \textit{Success weighted by Path Length (SPL)}.
Specifically, $\mathrm{SR}=\frac{1}{N}\sum_{i=1}^{N}\mathbb{I}[\|\mathbf{P}_i-\mathbf{G}_i\|_2<r_s]$,
where $\mathbf{P}_i$ and $\mathbf{G}_i$ denote the final and goal positions and $r_s$ is the success radius.
SPL is defined as $\mathrm{SPL}=\frac{1}{N}\sum_{i=1}^{N} S_i \frac{L_i^{*}}{\max(L_i,L_i^{*})}$,
where $L_i$ and $L_i^{*}$ denote the executed and geodesic path lengths, and $S_i\in\{0,1\}$ is the success indicator.

\subsection{Results of Evidence Arbitration for Decision}
\label{sec:exp_evidence}

This experiment evaluates how confidence-calibrated semantic evidence arbitration (Sec.~III-C) affects the quality of semantic inputs for downstream decisions. Rather than treating perception as an isolated classification task, we examine whether the generated semantic evidence enables \emph{correct semantic decisions} during autonomous exploration. Accordingly, we report SA, defined as whether the semantic category of the reached target matches the instruction.

Experiments are conducted in five open-set simulation environments, including both indoor and outdoor scenes. To provide a direct comparison with publicly available methods, we include representative open-source baselines (Qwen2.5-VL, GroundingDINO, and YOLO-World), together with a naive Qwen + GroundingDINO fusion and the proposed confidence-calibrated evidence arbitration that enforces confidence normalization and spatial consistency.
Results are summarized in Table~\ref{tab:perception_results}.
Scene-level reasoning alone exhibits noticeable instability across cluttered environments, while object-level detection benefits from precise localization but remains sensitive to occlusion and viewpoint changes. Naive fusion partially mitigates these issues but yields limited improvement.
In contrast, confidence-calibrated arbitration consistently achieves the highest semantic accuracy across all scenes. Compared with the strongest baseline (Qwen + GroundingDINO), our method improves average SA from 85.3\% to 90.1\%, yielding a +4.8 percentage-point gain.

These results indicate that the primary contribution of evidence arbitration lies not in improving raw recognition rates, but in filtering unreliable semantic cues before they influence decision making.
By stabilizing heterogeneous semantic evidence through confidence calibration and spatial consistency, the proposed mechanism provides more reliable decision inputs for subsequent subgoal selection and topological reasoning.

\begin{table}[t!]
\renewcommand\arraystretch{1.3}
\caption{Semantic Accuracy (\%) $\uparrow$ of Different Evidence Integration Strategies across Scenes. WH. means Warehouse}
\vspace{-8pt}
\centering
\setlength{\tabcolsep}{1.0mm}
\footnotesize
\begin{tabular}{lccccc}
\hline
\textbf{Method} & \textbf{Garden} & \textbf{Sidewalk} & \textbf{Road} & \textbf{WH. (a)} & \textbf{WH. (b)} \\
\hline
Qwen2.5-VL \cite{bai2023qwen} & 80.2 & 81.6 & 84.0 & 80.6 & 79.5 \\
GroundingDINO \cite{liu2023grounding} & 81.7 & 83.3 & 85.6 & 82.9 & 80.8 \\
YOLO-World \cite{Cheng2024YOLOWorld} & 83.1 & 84.4 & 85.9 & 83.5 & 82.2 \\
Qwen+GroundingDINO & 84.7 & 86.1 & 88.2 & 83.8 & 83.5 \\
Ours & \textbf{88.9} & \textbf{91.0} & \textbf{92.3} & \textbf{89.6} & \textbf{88.7} \\
\hline
\end{tabular}
\vspace{-6pt}
\label{tab:perception_results}
\end{table}

\begin{table}[t!]
\renewcommand\arraystretch{1.3}
\caption{GNSA (\%) $\uparrow$ of Different Decision Strategies across Scenes.}
\vspace{-8pt}
\centering
\setlength{\tabcolsep}{1.3mm}
\footnotesize
\begin{tabular}{lccccc}
\hline
\textbf{Method} & \textbf{Garden} & \textbf{Sidewalk} & \textbf{Road} & \textbf{WH. (a)} & \textbf{WH. (b)} \\
\hline
Bayesian Probability & 76.5 & 78.4 & 79.7 & 78.8 & 75.9 \\
GPT-4 \cite{achiam2023gpt} & 79.4 & 81.6 & 82.9 & 81.8 & 80.2 \\
VLFM \cite{yokoyama2024vlfm} & 81.1 & 83.3 & 84.7 & 83.5 & 80.8 \\
HOV-SG \cite{werby23hovsg} & 82.5 & 83.8 & 85.2 & 84.4 & 82.6 \\
Ours & \textbf{84.6} & \textbf{86.1} & \textbf{87.3} & \textbf{86.6} & \textbf{84.2} \\
\hline
\end{tabular}
\vspace{-17pt}
\label{tab:planning_results}
\end{table}

\subsection{Utility-Driven Subgoal Decision on Topological Memory}
\label{sec:exp_utility}

This experiment evaluates the effectiveness of the proposed subgoal decision mechanism on topological memory.
Rather than assessing low-level path execution, we focus on the quality of \emph{global decision making}: whether the system consistently selects semantically and spatially appropriate subgoals when multiple candidates exist. We therefore adopt GNSA as a proxy for decision optimality, complemented by OASR to verify execution feasibility.

We compare several subgoal decision strategies operating on the same semantic topological memory.
\emph{Bayesian Probability} ranks candidate nodes by fused semantic confidence.
\emph{GPT-4} estimates semantic relevance between node labels and the instruction using a language model.
\emph{Ours} applies the proposed semantic utility formulation that jointly considers semantic relevance, confidence, exploration potential, and traversal cost.
For broader comparison, we additionally include representative open-source methods (VLFM and HOV-SG), where only their subgoal decision components are applied while the exploration framework remains identical.
This setup isolates the impact of different decision criteria.

Results are summarized in Table~\ref{tab:planning_results}.
Confidence-only ranking exhibits limited decision stability, particularly in cluttered scenes where high-confidence nodes may incur high traversal cost.
Introducing language-model reasoning (GPT-4) consistently improves GNSA across environments, indicating better alignment between semantic inference and task instructions.
Our utility-driven strategy further improves decision quality by explicitly incorporating traversal cost and exploration value, resulting in more executable subgoal selections.
Compared with the strongest open-source baseline (HOV-SG), our method increases average GNSA from $83.7\%$ to $85.8\%$ (+2.1\%).

\begin{table}[t!]
\renewcommand\arraystretch{1.3}
\caption{Object Exploration Results (\%) in Simulation and Real-World Environments (20 trials per scene). Env. means Environment, Sim. means Simulation}
\vspace{-8pt}
\centering
\setlength{\tabcolsep}{1.8mm}
\footnotesize
\begin{tabular}{cll|ccc}
\hline
\textbf{Env.} & \textbf{Scene} & \textbf{Goal Object} & \textbf{OASR} $\uparrow$ & \textbf{SR} $\uparrow$ & \textbf{SPL} $\uparrow$ \\
\hline
\multirow{5}{*}{Sim.} 
 & Garden & \emph{Green Bin} & 79.3 & 55 & 33.6  \\
 & Sidewalk & \emph{Traffic Sign} & 80.9 & 60 & 40.5 \\
 & Road & \emph{Bus Station} & 84.2 & 75 & 46.3 \\
 & Warehouse (a) & \emph{Loading Cart} & 82.7 & 65 & 42.2 \\
 & Warehouse (b) & \emph{Extinguisher} & 79.5 & 50 & 29.7 \\
\hline
\multirow{5}{*}{Real} 
 & Office & \emph{Extinguisher} & 75.8 & 50 & 33.5 \\
 & Showroom & \emph{Quadruped Robot} & 77.2 & 55 & 35.4 \\
 & Laboratory & \emph{Black Chair} & 75.6 & 45 & 33.2 \\
 & Living Room & \emph{Bucket Shelf} & 71.9 & 40 & 26.8 \\
 & Garden & \emph{Carton Box} & 73.1 & 45 & 28.9 \\
\hline
\end{tabular}
\vspace{-17pt}
\label{tab:nav_results}
\end{table}

\begin{figure*}[t!]
\centering
\includegraphics[width=.99\textwidth]{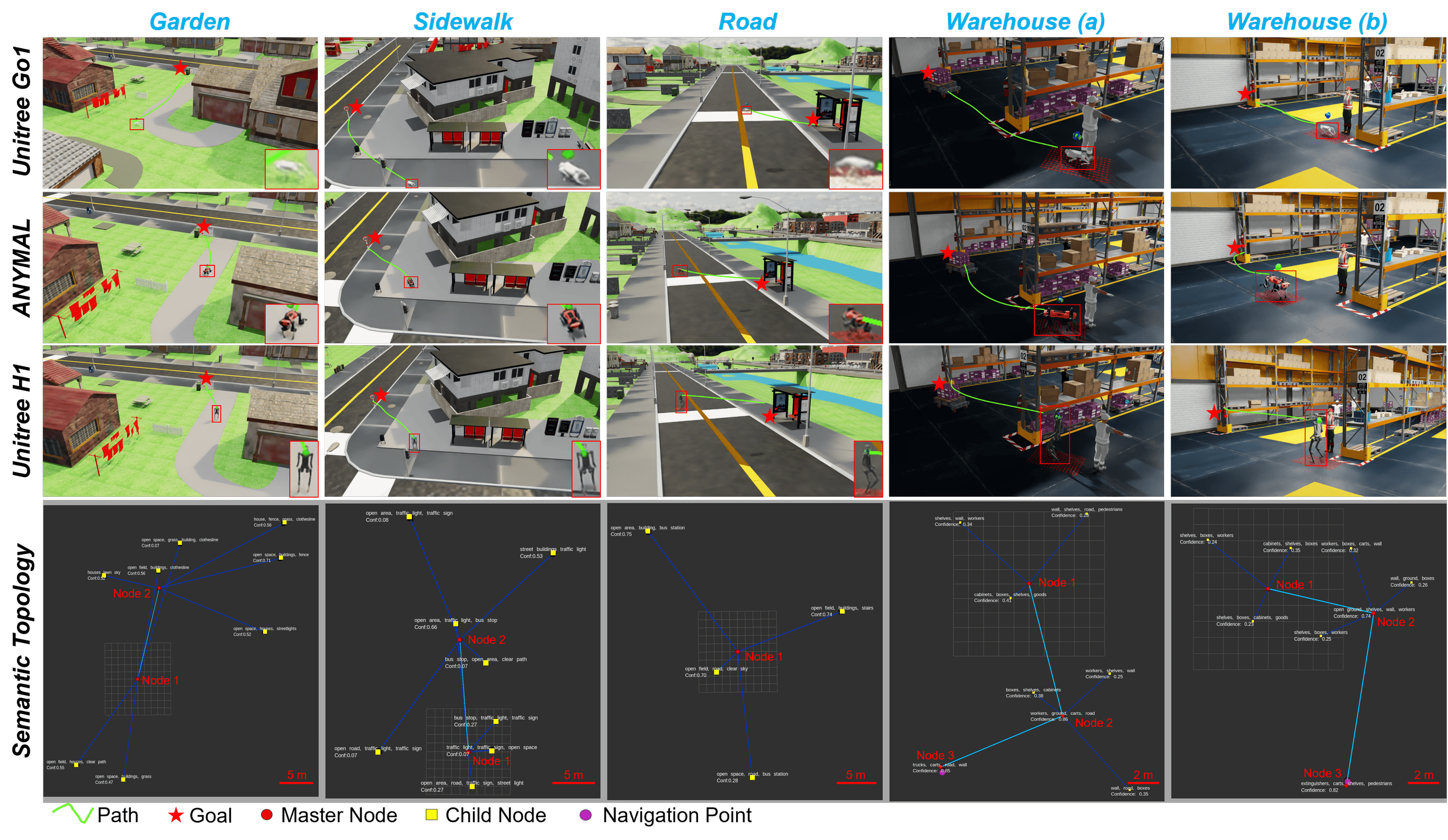}
\vspace{-12pt}
\caption{\textbf{Simulation results across five scenes with different robots.} We evaluate our exploration framework in \emph{Garden}, \emph{Sidewalk}, \emph{Road}, and two \emph{Warehouse} scenarios. Experiments are conducted using three different legged robot platforms, demonstrating the cross-platform adaptability.}
\label{simulation}
\vspace{-15pt}
\end{figure*}

\subsection{System-Level Object Exploration Experiments}
\label{sec:exp_system}

We evaluate our framework through system-level object exploration experiments in both simulated and real-world environments. All experiments are conducted without prior maps, and performance is measured using SR and SPL, reflecting task completion reliability and exploration efficiency.

\subsubsection{\textbf{Simulation Experiments}}
Fig.~\ref{simulation} illustrates representative exploration trajectories and semantic topological maps across five simulated environments. Quantitative results are summarized in Table~\ref{tab:nav_results}. The system shows stable performance in scenes with clear semantic cues and open geometry, achieving higher success rates and competitive SPL. In more cluttered environments, failures mainly stem from unstable semantic evidence or limited local reachability. Nevertheless, successful runs still produce trajectories close to geodesic paths, indicating coherent decision sequences with minimal backtracking.

\subsubsection{\textbf{Real-world Experiments}}
\begin{figure}[t]
\centering
\includegraphics[width=0.48\textwidth]{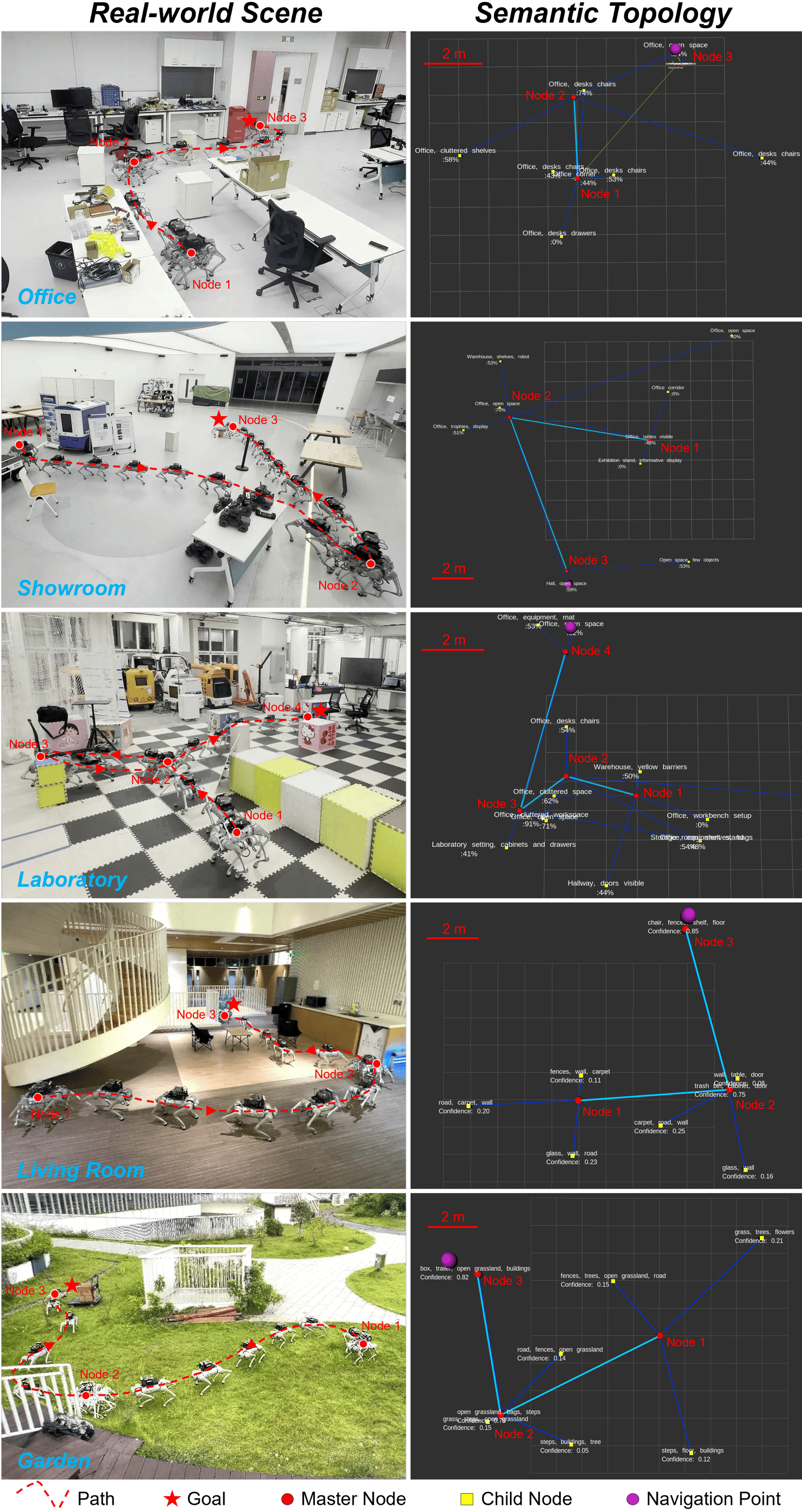}
\vspace{-20pt}
\caption{\textbf{Real-world results across five scenes.} 
Exploration trajectories and semantic topological maps are shown for Office, Showroom, Laboratory, Living Room, and Garden using Unitree Go1.}
\label{real-world}
\vspace{-16pt}
\end{figure}

We further deploy the system on a Unitree Go1 quadruped robot and evaluate it in five real-world environments. Fig.~\ref{real-world} shows representative robot trajectories and the corresponding semantic topological maps, where the robot incrementally expands semantic memory and converges toward task-relevant targets. Quantitative results are also reported in Table~\ref{tab:nav_results}. Compared to simulation, performance decreases due to perception noise, motion blur, and illumination variations; however, the framework consistently completes object exploration tasks across diverse indoor and outdoor settings, demonstrating the robustness of the proposed decision-driven exploration mechanism in real-world deployment.

\subsection{Runtime Efficiency Analysis}
\label{sec:exp_runtime}

Although the proposed framework incorporates both VLM and LLM inference, these models are not placed inside the high-frequency motion-control loop. Instead, the system follows a \emph{hierarchical, event-triggered} execution scheme: high-level semantic modules are invoked only when the robot reaches a stable viewpoint to generate semantic evidence, update the topological memory, and determine the next subgoal. Once a subgoal is selected, execution is handled independently by the local obstacle-aware planner and the low-level locomotion policy at real-time rates. Consequently, large-model latency mainly affects the \emph{decision refresh interval} rather than the stability of motion control.

Table~\ref{tab:runtime_analysis} summarizes the update rate and latency of each module. RGB-D observations and robot states are streamed continuously at 30~Hz, whereas semantic reasoning modules are invoked \emph{on demand}. In practice, Qwen-VL and GroundingDINO run in parallel, resulting in an overall latency of about 3-3.5~s for semantic perception and topology construction at each viewpoint. By contrast, the execution layer operates continuously, with trajectory tracking at 12~Hz and motion control at 50~Hz. This decoupled design makes the system more sensitive to low-level control frequency than to high-level semantic inference latency.

\begin{table}[t]
\renewcommand\arraystretch{1.3}
\caption{Runtime analysis of the exploration system.}
\vspace{-8pt}
\centering
\setlength{\tabcolsep}{1.0mm}
\footnotesize
\begin{tabular}{lccc}
\hline
\textbf{Module} & \textbf{Update Rate} & \textbf{Triggering} & \textbf{Latency} \\
\hline
RGB-D / Robot State & 30 Hz & Continuous stream & -  \\
Qwen-VL Reasoning & On demand & Per viewpoint & $\sim$2.5 s \\
GroundingDINO & On demand & Per viewpoint & $\sim$1.2 s  \\
Evidence Arbitration & On demand & After detection & $\sim$15 ms \\
Topology Update & On demand & After arbitration & -  \\
Utility Decision & On demand & Goal selection & $\sim$2.0 s  \\
Trajectory Tracking & 12 Hz & During execution & Real-time  \\
Motion Policy & 50 Hz & During execution & Real-time  \\
\hline
\end{tabular}
\vspace{-19pt}
\label{tab:runtime_analysis}
\end{table}

\subsection{Ablation Studies}
\label{sec:exp_ablation}

We conduct a system-level ablation study to evaluate the impact of the proposed decision mechanisms, including semantic evidence arbitration and utility-driven subgoal decision. Performance is measured using SR and SPL, reflecting exploration reliability and efficiency.

Results are summarized in Table~\ref{tab:ablation}. The baseline system relies on scene-level perception and confidence-based ranking for subgoal selection, achieving only $35\%$ SR and $23.5\%$ SPL due to unstable semantic observations.
Introducing semantic evidence arbitration improves SR to $45\%$ and SPL to $30.8\%$, indicating that calibrating heterogeneous semantic cues provides more reliable decision inputs for exploration.
Further incorporating the proposed utility-driven subgoal decision increases SR to $55\%$ and SPL to $34.2\%$. By jointly considering semantic relevance, confidence, exploration potential, and travel cost, the robot selects more executable subgoals and produces more efficient trajectories.

Overall, the results show that evidence arbitration stabilizes semantic perception, while utility-driven decision further improves global exploration planning.

\begin{table}[t]
\renewcommand\arraystretch{1.3}
\caption{Ablation Results of Core Decision Mechanisms on SR and SPL (\%). Arb. means Arbitration}
\vspace{-5pt}
\centering
\setlength{\tabcolsep}{2.5mm}
\footnotesize
\begin{tabular}{lcc|cc}
\hline
\textbf{Method} & \textbf{\makecell{Evidence\\Arbitration}} & \textbf{\makecell{Utility\\Decision}} & \textbf{SR} $\uparrow$  & \textbf{SPL} $\uparrow$ \\
\hline
Baseline & \textcolor[HTML]{B22222}{\ding{55}} & \textcolor[HTML]{B22222}{\ding{55}} & 35 & 23.5 \\
+ Evidence Arb. & \textcolor[HTML]{228B22}{\ding{51}} & \textcolor[HTML]{B22222}{\ding{55}} & 45 & 30.8 \\
Full Used (Ours) & \textcolor[HTML]{228B22}{\ding{51}} & \textcolor[HTML]{228B22}{\ding{51}} & \textbf{55} & \textbf{34.2} \\
\hline
\end{tabular}
\label{tab:ablation}
\vspace{-19pt}
\end{table}



\section{Conclusion}
\label{sec:discussion_conclusion}
This work highlights decision mechanisms in semantic exploration and presents a decision-driven semantic object exploration framework for legged robots, formulating open-world exploration as a subgoal selection process constrained by calibrated semantic evidence.
Experiments demonstrate that compact semantic topology is sufficient for goal-directed exploration, provided that evidence arbitration and utility balancing remain stable.
Current validation is conducted in moderately scaled and mostly static environments. System performance still depends on perceptual stability; accumulated semantic noise in long-horizon or highly dynamic scenarios may affect decision reliability.
Overall, this work offers a structured decision perspective for semantic exploration under resource constraints and provides a foundation for deployment in more complex environments.
Future work will incorporate temporal consistency modeling and explicit uncertainty propagation, and further evaluate generalization in more complex real-world settings.


\bibliographystyle{IEEEtran}
\bibliography{ref.bib} 

\end{document}